\documentclass[11pt,letterpaper]{article}
\usepackage{naaclhlt2016}
\usepackage{times}
\usepackage{latexsym}
\usepackage{graphicx}

\naaclfinalcopy 
\def\naaclpaperid{***} 

\title{The Effect of Negators, Modals, and Degree Adverbs\\ on Sentiment Composition}

\author{Svetlana Kiritchenko \and Saif M. Mohammad\\
	    National Research Council Canada\\
	    {\tt \small \{svetlana.kiritchenko,saif.mohammad\}@nrc-cnrc.gc.ca}
}

\date{}

\begin{document}

\maketitle

\newcommand{\measure}[2]{$\rm \it #1_{{\rm \it #2}}$}
\newcommand{\score}[1]{$\rm \it score({\rm \it #1})$}

\begin{abstract}
Negators, modals, and degree adverbs can significantly affect the sentiment of the words they modify. 
Often, their impact is modeled with simple heuristics; although, recent work has shown that such heuristics do not capture the true sentiment of multi-word phrases. 
We created a dataset of phrases that include various negators, modals, and degree adverbs, as well as their combinations.
Both the phrases and their constituent content words were annotated with real-valued scores of sentiment association. 
Using phrasal terms in the created dataset, we analyze the impact of individual modifiers and the average effect of the groups of modifiers on overall sentiment.
We find that the effect of modifiers varies substantially among the members of the same group. 
Furthermore, each individual modifier can affect sentiment words in different ways. 
Therefore, solutions based on statistical learning seem more promising than fixed hand-crafted rules on the task of automatic sentiment prediction.  
\end{abstract}

\section{Introduction}

Sentiment associations are commonly captured in sentiment lexicons---lists of associated word--sentiment pairs (optionally with a score indicating the degree of association). 
They are mostly used in sentiment analysis \cite{Semeval2014task4,Rosenthal-EtAl:2015:SemEval}, but are also beneficial in 
stance detection \cite{stance-lrec,MohammadSK16}, 
\cite{hartner2013lingering,kleres2011emotions,Mohammad2012},
detecting personality traits \cite{grijalva2014gender,COIN:COIN12024}, 
and other applications. 

Manually created sentiment lexicons are especially useful because they tend to be more accurate than automatically generated ones; 
they can be used to automatically generate large high-coverage lexicons \cite{tang2014building,Esuli06}; 
they can be used to evaluate different methods of automatically creating sentiment lexicons; 
and they can be used for linguistic analysis such as examining how modifiers (negators, modals, degree adverbs, etc.) impact overall sentiment. 
However, most existing manually created sentiment lexicons tend to provide only lists of positive and negative words with very coarse levels of sentiment
\cite{Stone66,Wilson05,MohammadT13}.  
The coarse-grained distinctions may be less useful in downstream applications than having access to fine-grained (real-valued) sentiment association scores. 

Manually created sentiment lexicons usually include only single words. 
Yet, the sentiment of a phrase can differ markedly from the sentiment of its constituent words. 
Sentiment composition is the determining of sentiment of a multi-word linguistic unit, such as a phrase or a sentence, from its constituents. 
Lexicons that include sentiment associations for phrases as well as for their constituents are useful in studying sentiment composition. 
We refer to them as {\it sentiment composition lexicons (SCLs)}. 

We created a sentiment composition lexicon for phrases formed with negators (such as \textit{no} and \textit{cannot}), modals (such as \textit{would have been} and \textit{could}), degree adverbs (such as \textit{quite} and \textit{less}), and their combinations. 
Both the phrases and their constituent content words were manually annotated with real-valued scores of sentiment association using a technique known as Best--Worst Scaling, which provides reliable annotations. 
We refer to the resulting lexicon as {\it Sentiment Composition Lexicon for Negators, Modals, and Degree Adverbs (SCL-NMA)}. 
The lexicon is also known as {\it SemEval-2016 General English Sentiment Modifiers Lexicon}.\footnote{This lexicon was first introduced in  \cite{maxdiff-naacl2016} where we investigated the applicability and reliability of the Best--Worst Scaling annotation technique in capturing word--sentiment associations. In this paper, we provide further details on the creation of the lexicon and present analysis of how negators, modals, and degree adverbs impact the sentiment of the words they modify.} 

We calculate the minimum difference in sentiment scores of two terms that is perceptible to native speakers of a language. 
For sentiment scores between -1 and 1, we show that the perceptible difference is about 0.07 for English speakers.
Knowing the least perceptible difference helps interpret the impact of sentiment composition. For example, we can
determine whether a modifier significantly impacts the sentiment of the word it composes with by calculating the difference in sentiment scores between the combined phrase and
the constituent, and checking whether this difference is greater than the least perceptible difference.

We use the phrasal terms in the created lexicon to analyze the impact of common modifiers on the sentiment of the terms they modify. 
We measure the effect of individual modifiers as well as the average effect of the groups of modifiers on overall sentiment.
We show that the sentiment of a negated expression (such as {\it not w}) on the [-1,1] scale is on average 0.926 points less than the sentiment of the modified term {\it w}, if the {\it w} is positive. 
However, the sentiment of the negated expression is on average 0.791 points higher than {\it w}, if the {\it w} is negative. 
Similar analysis for modals and degree adverbs shows that they impact sentiment less dramatically than negators. 
Furthermore, the impact of modifiers substantially varies even within a group, e.g., the average change in sentiment score brought by the negator `{\it will not be}' is 0.41 larger than the change introduced by the negator `{\it never}'. 
Likewise, each individual modifier can affect sentiment words in different ways. 
As a result, in automatic sentiment prediction solutions based on statistical learning seem more promising than fixed hand-crafted rules.  

In related work (not described here), we also created a sentiment composition lexicon for another challenging category of phrases---phrases that include at least one positive word and at least one negative word \cite{mixedpol-naacl2016}. 
We call such phrases opposing polarity phrases. 
Both lexicons have been used as official test sets in SemEval-2016 Task 7 `Determining Sentiment Intensity of English and Arabic Phrases' \cite{SemEval2016Task7}.\footnote{http://alt.qcri.org/semeval2016/task7/}  
The lexicons are made freely available to the research community.\footnote{http://www.saifmohammad.com/WebPages/SCL.html}

\section{Related Work}

\noindent {\bf Sentiment Lexicons: }
There exist a number of manually created lexicons that provide lists of positive and negative words, for example, General Inquirer \cite{Stone66}, Hu and Liu Lexicon \cite{Hu04}, and NRC Emotion Lexicon \cite{MohammadT13}. 
Only a few manually created lexicons provide real-valued scores of sentiment association \cite{bradley1999affective,warriner2013norms,dodds2011temporal}.
None of these lexicons, however, contain multi-word phrases. 
Manually created sentiment lexicons can be used to automatically generate larger sentiment lexicons using
semi-supervised techniques \cite{Esuli06,TurneyL03,MohammadSemEval2013,de2013good,tang2014building}. (See \newcite{SentimentEmotionSurvey2015} for a survey on manually created and automatically generated affect resources.)

Automatically generated lexicons often have real-valued sentiment association scores, are larger in scale, and can easily be collected for a specific domain; therefore, they were found to be more beneficial in downstream applications, such as sentence-level sentiment prediction \cite{Kiritchenko2014}.  
However, any analysis of the relationship between the sentiment of a phrase and its constituents is less reliable when made from an automatically generated resource as opposed to when made from a manually created resource (as automatically generated resources are less accurate). 
In this work, we provide an extensive analysis of the impact of different modifiers on sentiment based on reliable fine-grained manual annotations.

\newpage
\noindent {\bf Contextual Valence Shifters: }
Negators, modals, and degree adverbs impact the sentiment of the word or phrase
they modify and are commonly referred to as contextual valence shifters
\cite{Polanyi04,Kennedy05,Jia09,Wiegand10,LapponiRO12}.
Conventionally, the impact of contextual valence shifters is captured by simple heuristics. 
For example, negation is often handled by reversing the polarities of the sentiment words in the scope of negation \cite{Polanyi04,Kennedy05,ChoiC08} or by shifting the sentiment score of a term in a negated context towards the opposite polarity by a fixed amount \cite{Taboada_2011}.
However, such heuristics do not adequately capture the true sentiment of multi-word expressions \cite{Zhu2014}. 
\newcite{liu2009review} relax the assumption of a fixed shifting margin and estimate these margins for each modifier separately from data. 
\newcite{Kiritchenko2014}, on the other hand, estimate the impact of negation on each individual sentiment word through a corpus-based statistical method. 
\newcite{ruppenhoferordering} automatically rank English adverbs by their intensifying or diminishing effect on adjectives using ratings metadata from product reviews.  

\noindent \textbf{Annotation techniques:} A widely used method of annotation for obtaining numerical scores is the {\it rating scale} method---where 
one is asked to rate an item on a five-, ten-, or hundred-point scale. 
While easy to understand, rating items on a scale is not natural for people. 
It is hard for annotators to remain consistent when annotating a large number of items.
Also, respondents often use just a limited part of the scale reducing the discrimination among items \cite{Cohen_2003}.
To obtain reliable annotations, the rating scale methods require a high number of responses, typically 15 to 20 \cite{warriner2013norms,grahamaccurate}.
A more natural annotation task for humans is to compare items (e.g., whether one word is more positive than the other). 
Most commonly, the items are compared in pairs 
\cite{thurstone1927law,david1963method}.
In this work, we use {\it Best--Worst Scaling}---a technique that exploits the comparative approach to annotation while keeping the number of required annotations small (Section~\ref{maxdiff}).
It has been shown to produce reliable annotations of terms by sentiment \cite{maxdiff-naacl2016}.

\section{Creating SCL-NMA}
\label{lexicon-description}

\begin{table}[t!]
\begin{center}
\small{
\begin{tabular}{lr}
\hline {\bf Term} & {\bf Sentiment}\\
 & {\bf score}\\ \hline
favor & 0.653\\
would be very easy & 0.431\\
did not harm & 0.194\\
increasingly difficult & -0.583\\
severe & -0.833\\
\hline
\end{tabular}
}
\caption{\label{tab:lex-examples} {Example entries with real-valued sentiment scores from SCL-NMA.}
}
\end{center}
\vspace*{-3mm}
\end{table}

We now describe the term selection process and the Best--Worst Scaling annotation technique used to create the Sentiment Composition Lexicon for Negators, Modals, and Degree Adverbs.
Table~\ref{tab:lex-examples} shows a few example entries from the lexicon. 
We also describe how we calculated the minimum difference in sentiment scores of two terms that is perceptible to native speakers of a language.

\subsection{Term Selection}
\label{term-selection}

General Inquirer \cite{Stone66} provides a list of 1,621 positive and negative words from Osgood's seminal study on word meaning \cite{Osgood57}. 
These are words commonly used in everyday English.
We include all of these words.
In addition, we include 1,586 high-frequency phrases formed by the Osgood words in combination with simple negators such as {\it no}, {\it don't}, and {\it never}, modals such as {\it can}, {\it might}, and {\it should}, or degree adverbs such as {\it very} and {\it fairly}.\footnote{The complete lists of negators, modals, and degree adverbs used to create this dataset are available at http://www.saifmohammad.com/WebPages/SCL.html\#NMA.}
The eligible adverbs are chosen manually from adverbs frequently occurring in the British National Corpus (BNC)\footnote{The British National Corpus, version 3 (BNC XML Edition). 2007. Distributed by Oxford University Computing Services on behalf of the BNC Consortium. http://www.natcorp.ox.ac.uk/}.
Each phrase includes at least one modal, one negator, or one adverb; a phrase can include several modifiers (e.g., \textit{would be very happy}). 
The modifiers and the phrases are chosen in such a way that the full set includes several phrases for each Osgood sentiment word and includes several phrases for each modifier.  
In total, sixty-four different (single or multi-word) modifiers are selected. 
The final list contains 3,207 terms. 

\subsection{Best--Worst Scaling}
\label{maxdiff}

Best--Worst Scaling (BWS), also sometimes referred to as Maximum Difference Scaling (MaxDiff), is an annotation scheme that exploits the comparative approach to annotation \cite{Louviere_1990,Cohen_2003,Louviere2015}. 
Annotators are given four items (4-tuple) and asked which item is the Best (highest in terms of the property of interest) and which is the Worst (least in terms of the property of interest). 
These annotations can then be converted into real-valued scores and also a ranking  of items as per their association with the property of interest through a simple counting procedure: For each item, its score is calculated as the percentage of times the item was chosen as the Best minus the percentage of times the item was chosen as the Worst
\cite{Orme_2009,flynn2014}.
The scores range from -1 to 1. 
Further details on Best--Worst Scaling and its application to the task of sentiment annotation can be found in \cite{maxdiff-naacl2016}.

\subsection{Annotation process} 
The complete list of 3,207 terms was randomly sampled (with replacement) to create  6,414 (2 x 3,207) 4-tuples that satisfy the following criteria:
\begin{enumerate}
\vspace*{-2mm}
\item no two 4-tuples have the same four terms; 
\vspace*{-2mm}
\item no two terms within a 4-tuple are identical; 
\vspace*{-2mm}
\item each term in the term list appears approximately in the same number of 4-tuples; 
\vspace*{-2mm}
\item each pair of terms appears approximately in the same number of 4-tuples. 
\end{enumerate}
\noindent Next, the set of 4-tuples was annotated through a crowdsourcing platform, CrowdFlower. 
The annotators were presented with four terms (single words and multi-word phrases) at a time, and asked which term is the most positive (or least negative) and which is the most negative (or least positive).\footnote{The full set of instructions to annotators is available at  http://www.saifmohammad.com/WebPages/SCL.html\#NMA.} 
Each 4-tuple was annotated by ten respondents. 
We determined accuracy of every annotator on a small set of check questions labeled by the authors of this paper.  
We discarded all annotations provided by an annotator if their accuracy on these check questions was less than 70\%.

\subsection{Quality of Annotations}
\label{analysis-maxdiff}

Let {\it majority answer} refer to the option chosen most often for a question.
80\% of the responses to the Best--Worst questions matched the majority answer. 

We also tested the reliability of the aggregated scores by randomly dividing the sets of ten responses to each question into two halves and comparing the rankings obtained from these two groups of responses.
The Spearman rank correlation coefficient between the two sets of rankings was found to be 0.98. (The Pearson correlation coefficient between the two sets of sentiment scores was also 0.98.)
Thus, even though annotators might disagree about answers to individual questions, the aggregated scores produced by applying the counting procedure 
on the Best--Worst annotations are remarkably reliable at ranking terms by sentiment.

\subsection{Least Perceptible Difference in Sentiment}

In psychophysics, there is a notion of {\it least perceptible difference} (aka {\it just-noticeable difference})---the amount by which something that can be measured (e.g., weight or sound intensity) needs to be changed in order for the difference to be noticeable by a human \cite{fechner1966elements}.
Analogously, we can measure the least perceptible difference in sentiment. 
If two words have close to identical sentiment associations, then
it is expected that native speakers will choose each of the words about the same number of times when forced to pick
a word that is more positive. However, as the difference in sentiment starts getting larger,
the frequency with which the two terms are chosen as most positive begins to diverge.
At one point, the frequencies diverge so much that we can say with high confidence that the two terms do not have the same sentiment associations.  
The average of this minimum difference in sentiment score is the least perceptible difference for sentiment. 

\begin{figure}[t]
\centering
\includegraphics[width=3in]{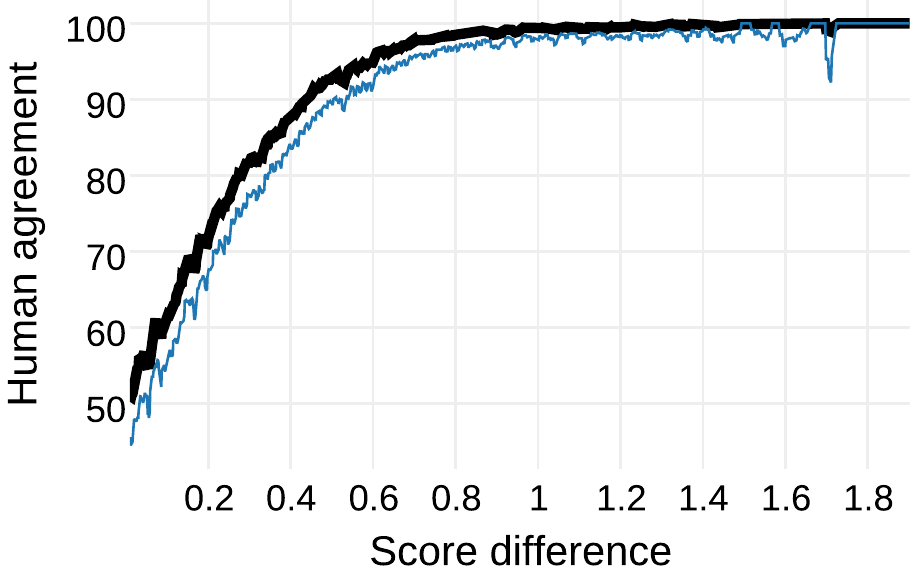}
\caption{\small Human agreement on annotating term $w_{1}$ as more positive than term $w_{2}$ for pairs with difference in scores $d$ = \score{w_{1}} - \score{w_{2}}.  
The x-axis represents $d$.
The y-axis plots the avg.\@ percentage of human annotations that judge term $w_{1}$ as more positive than term $w_{2}$ (thick line) and the corresponding 99.9\%-confidence lower bound (thin blue line).
}
\label{human-agreement}
\vspace*{-5mm}
\end{figure}

To calculate the least perceptible difference, we first build a plot of the relationship between 'difference in the sentiment scores between two terms' and `agreement among annotators' when asked which term is more positive.  
For each term pair $w_1$ and $w_2$ such that $d = $ \score{w_{1}} $-$ \score{w_{2}} $\geq$ 0, we count the number of Best--Worst annotations from which we can infer that $w_{1}$ is more positive than $w_{2}$  and divide this number by the total number of annotations from which we can infer either that $w_{1}$ is more positive than $w_{2}$ or that $w_{2}$ is more positive than $w_{1}$.
(We can infer that $w_{1}$ is more positive than $w_{2}$ if in a 4-tuple that has both $w_1$ and $w_2$ the annotator selected $w_{1}$ as the most positive or $w_{2}$ as the least positive. The case for $w_{2}$ being more positive than $w_{1}$ is similar.)
This ratio is the human agreement for $w_1$ being more positive than $w_2$. 
To get more reliable estimates, we average the human agreement for all pairs of terms whose sentiment differs by $d \pm 0.01$. 
Figure~\ref{human-agreement} shows the resulting average human agreement. 
The thin blue line in the Figure depicts the 99.9\%-confidence lower bounds on the  agreement. 
The least perceptible difference is the point starting at which the lower bound consistently exceeds 50\% threshold (i.e., the point starting at which we observe with 99.9\% confidence that the human agreement is higher than chance).
The least perceptible difference when calculated from SCL-NMA is 0.069. 
In the next section, we use the least perceptible difference to determine whether a modifier significantly impacts the sentiment of the word it composes with.

\section{Impact of Negators, Modals, and Degree Adverbs on Sentiment}
\label{analysis-modifiers}

\begin{table*}[t!]
\caption{\label{tab:modifiers} The impact of different modifier groups on sentiment. 
`Avg. diff.' is the average difference between the score of {\it mod w} and {\it w}.
`\# pairs` is the number of pairs (of {\it w} and {\it mod w})
used to determine the average.  
`\# score $\uparrow$ ($\downarrow$)` indicates the number of phrases for which \score{mod\ w} is greater (less) than   \score{w} by at least 0.069 (the perceptible difference).
}
\vspace*{-4mm}
\begin{center}
\resizebox{\textwidth}{!}{
\begin{tabular}{l rrrr rrrr}
\hline 
{\bf Modifier Group} & \multicolumn{4} {c }{\bf On positive words} & \multicolumn{4}{c}{\bf On negative words}\\
& Avg. diff. & \# pairs & \# score $\uparrow$ & \# score $\downarrow$ &  Avg. diff. & \# pairs & \# score $\uparrow$ & \# score $\downarrow$\\
\hline
negators & -0.926 & 265 & 1 & 264 & 0.791 & 71 & 71 & 0 \\
modals & -0.317 & 258 & 9 & 231 & 0.238 & 72 & 54 & 8 \\
degree adverbs (abs.\@ diff.) & 0.201 & 435 & 106 & 212 & 0.166 & 163 & 42 & 68\\
\hline
\end{tabular}
}
\end{center}
\end{table*}

SCL-NMA contains many phrases formed by different types of modifiers---negators, modals, and degree adverbs.
Thus, this lexicon is a good resource for studying the impact of these types of modifiers on sentiment.
In the following, we compare the sentiment score of single-word term $w$ with the sentiment score for phrase {\it mod w}, 
where {\it mod} is a modifier from a particular group (negator, modal, or degree adverb). 
Table~\ref{tab:modifiers} shows the average effect of different modifier groups on sentiment.
The columns show the average change in sentiment score between $w$ and {\it mod w}, the number of pairs (of $w$ and {\it mod w}) used to determine the average, the number of phrases {\it mod w} whose sentiment score is greater ($\uparrow$) or less ($\downarrow$) than the score of $w$ by at least 0.069 (the least perceptible difference). 
Since the impact of modifiers can be different depending on the sentiment of the modified word $w$,
we present separate analyses for when $w$ is positive and when $w$ is negative.
For the analysis in this section only, a word is considered positive if it has a sentiment score greater than or equal to 0.3, and 
considered negative if its sentiment score is less than or equal to -0.3.\footnote{This threshold is somewhat arbitrary, and is chosen to discard neutral terms from the analysis, whose sentiment tends not to change much with these modifiers.}

Observe that the most change in sentiment is caused by negation; it consistently decreases the scores of positive words, and increases the scores of negative words.
The average score difference is substantial for both positive words (0.926 points) and negative words (0.791 points).
Modals also tend to decrease the scores of positive words, and increase the scores of negative words, though to a much smaller extent than negators.
As with negators, modals affect positive words more strongly than they do negative words.
Degree adverbs show less consistency than negators and modals; they can both heighten or lower the sentiment of a word.
Moreover, the same adverb can behave differently with different words from the same sentiment group (positive or negative).
Therefore, we report the average \textit{absolute} differences in scores for this modifier group.
These average differences are substantially smaller than the ones reported for modals and negators; the effect of degree adverbs is minor.
Besides, in contrast to modals and negators, for a large percentage of degree adverb phrases (for 35\% of the positive-word phrases and for 37\% of the negative-word phrases), the sentiment scores do not differ from the scores for the corresponding single words by the least perceptible difference (0.069 points). 
In the subsections below, we further examine the impact of each modifier category on the sentiment of its scope. 
Also, we provide rankings of different negators, modals, and degree adverbs as per the average change in sentiment score between $w$ and {\it mod w}. 
This would allow linguists and other researchers to better understand the behavior of different modifiers.

\subsection{Negation}

\begin{figure}[!t]
\centering
\includegraphics[width=3.1in]{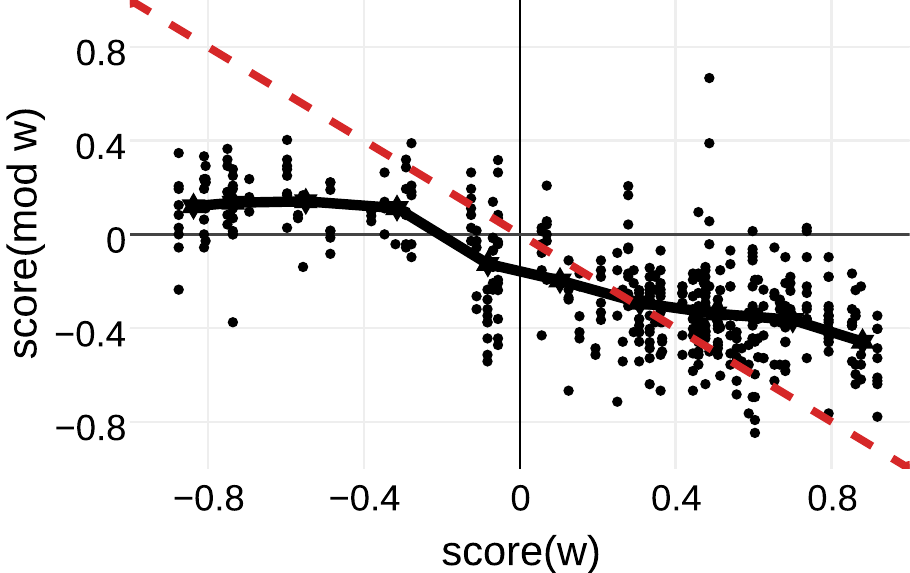}
\caption{\small The impact of \textbf{negators} on sentiment. 
The x-axis is \score{w}, the sentiment score of a term $w$; the y-axis is \score{mod\ w}, the sentiment score of a term $w$ preceded by a negator.
Each dot corresponds to one phrase \textit{mod w}.
The black line shows an average effect of the negators group.
The dashed red line shows the reversing polarity hypothesis \score{mod\ w}$=-$\score{w}.
}
\label{fig-negation}
\end{figure}

There exist two common approaches to incorporate the impact of negation in automatic systems: (1) {\it reversing polarity hypothesis}, where the sentiment score of a word `\score{w}' is replaced with `$-$\score{w}'; and (2)  \textit{shifting hypothesis}, where the sentiment score of a word `\score{w}' is reduced by a fixed amount: `\score{w}$-{\rm \it sign}($\score{w}$) \times b$'. 
We will show that neither hypothesis accurately captures the impact of negation.
We will also present an analysis of the overall impact of negation and the impact of individual negators (aka negation triggers).
 
In our dataset, the negators are formed by `no' negation words like \textit{no}, \textit{not}, \textit{never}, and \textit{nothing} in combination with auxiliary and modal verbs.
Figure~\ref{fig-negation} shows the overall impact of negation on sentiment of single words.
Each dot in this figure corresponds to one negated phrase `\textit{negator w}'.
The x-axis corresponds to \score{w} (the sentiment score of a word $w$); the y-axis is \score{mod\ w} (the sentiment score of a word $w$ preceded by a negator).
The black line shows an average effect of negation.
The dashed red line shows the reversing polarity hypothesis: \score{mod\ w}$=-$\score{w}.
Observe that on average negators tend to substantially downshift the sentiment of positive words turning them into negative expressions.
On the other hand, the scores of negative terms increase, but to a smaller extent than the scores of positive words.
Words with high absolute sentiment values tend to experience the greatest shift.
This is true for both positive and negative words.
However, this shift is substantially smaller than is proposed by the reversing polarity hypothesis.
Overall, the reversing polarity hypothesis fit is rather poor.
The shifting hypothesis does not explain the data either. 
Another observation is that words with similar sentiment scores can form negated phrases with very different sentiment scores (appearing as columns of dots in the graph).
This is mostly due to the effect of different negators.
However, the same negator can sometimes have different effect on words with similar sentiment.
For example, the three words \textit{easy}, \textit{good}, and \textit{better} all have similar sentiment scores: \score{easy} = 0.598, \score{good} = 0.556, \score{better} = 0.486. 
Yet, the corresponding negated phrases formed with the same negator \textit{never} range from negative (\score{never\ good} = $-$~0.542), to slightly negative (\score{never\ easy} = $-$~0.112), to positive (\score{never\ better} = 0.666). 

Next, we investigate the effect of individual negators.
Table~\ref{tab:negators} shows the impact of negation triggered by different negators.
The majority of the negation triggers have a large effect on both positive and negative words; the absolute difference in scores between a negated phrase and the corresponding sentiment word is 0.8-1.0 points on positive words and 0.7-0.9 points on negative words.
The greatest shift in sentiment on positive words was observed for the modifier \textit{will not be}, and on negative words for modifier \textit{will not}.
The weakest effect is caused by \textit{may not}, \textit{nothing}, and \textit{never}.
Verb tenses seem not to affect the behavior of negators significantly.
For example, the average change in sentiment caused by \textit{not} and by \textit{was not} differs only by 0.03-0.05 points.
The modal verbs \textit{will} and \textit{can} form strong negation phrases \textit{will not}, \textit{will not be}, and \textit{cannot} that showed the most change in sentiment.
Other modal verbs, such as \textit{could}, \textit{would}, and \textit{may}, form negation phrases with smaller effect on sentiment.

\setlength{\tabcolsep}{3pt}

\begin{table}[t]
\caption{\label{tab:negators} The impact of \textbf{negators} on sentiment. 
}
\begin{center}
\small{
\begin{tabular}{lrrrr}
\hline 
{\bf Modifier} & {\bf Avg. diff.} & {\bf \# pairs} & {\bf \# score $\uparrow$} & {\bf \# score $\downarrow$}\\
\hline
\multicolumn{2}{l}{\bf On positive words}\\
will not be & -1.066 & 9 & 0 & 9 \\
cannot & -1.030 & 12 & 0 & 12 \\
did not & -0.978 & 13 & 0 & 13 \\
not very & -0.961 & 14 & 0 & 14 \\
not & -0.959 & 45 & 0 & 45 \\
no & -0.948 & 29 & 0 & 29 \\
was no & -0.939 & 14 & 0 & 14 \\
will not & -0.935 & 11 & 0 & 11  \\
was not & -0.928 & 29 & 0 & 29  \\
have no & -0.917 & 8 & 0 & 8 \\
does not & -0.907 & 13 & 0 & 13  \\
could not & -0.893 & 12 & 0 & 12  \\
would not & -0.869 & 11 & 0 & 11\\
had no & -0.862 & 7 & 0 & 7  \\
would not be & -0.848 & 15 & 0 & 15  \\
may not & -0.758 & 6 & 0 & 6  \\
nothing & -0.755 & 6 & 0 & 6 \\
never & -0.650 & 7 & 1 & 6 \\[4pt]
\multicolumn{2}{l}{\bf On negative words}\\
will not & 0.878 & 5 & 5 & 0\\
does not & 0.823 & 5 & 5 & 0\\
was not & 0.786 & 10 & 10 & 0\\
no & 0.768 & 8 & 8 & 0\\
not & 0.735 & 14 & 14 & 0\\
\hline
\end{tabular}
}
\end{center}
\end{table}

\subsection{Modals}

\begin{figure}[!t]
\centering
\includegraphics[width=3.1in]{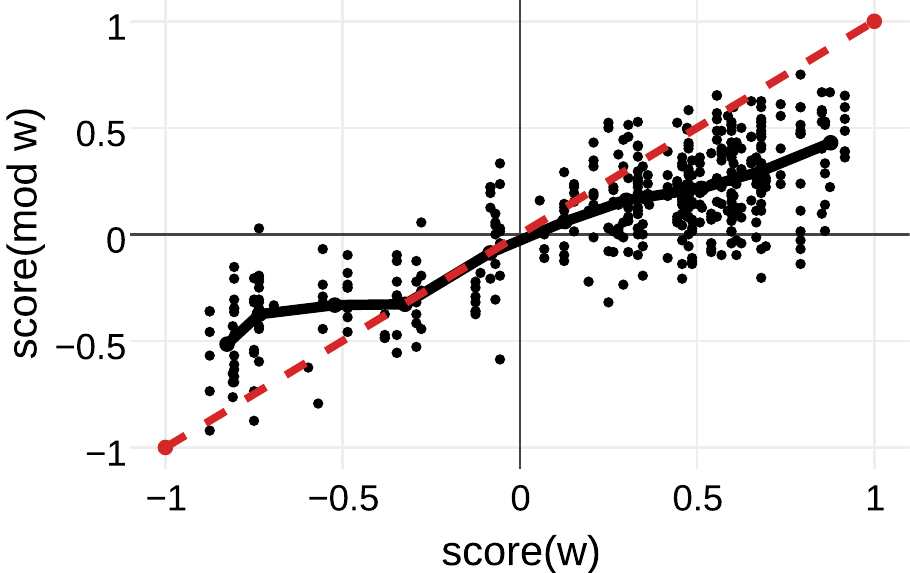}
\caption{The impact of \textbf{modals} on sentiment. 
The x-axis is \score{w}, the sentiment score of a term $w$; the y-axis is \score{mod\ w}, the sentiment score of a term $w$ preceded by a modal verb.
Each dot corresponds to one phrase \textit{mod w}.
The black line shows an average effect of the modals group.
The dashed red line shows the function \score{mod\ w} = \score{w}.
}
\label{fig-modals}
\end{figure}

\setlength{\tabcolsep}{2pt}

\begin{table}[t]
\caption{\label{tab:modals} The impact of \textbf{modals} on sentiment.}
\begin{center}
\small{
\begin{tabular}{lrrrr}
\hline  
{\bf Modifier} & {\bf Avg. diff.} & {\bf \# pairs} & {\bf \# score $\uparrow$} & {\bf \# score $\downarrow$}\\
\hline
\multicolumn{2}{l}{\bf On positive words}\\
would have been & -0.491 & 12 & 0 & 12 \\
could & -0.390 & 16 & 1 & 15 \\
might & -0.387 & 14 & 0 & 14 \\
may & -0.384 & 14 & 1 & 13 \\
should be & -0.365 & 22 & 0 & 22 \\
could be & -0.342 & 14 & 0 & 13 \\
must & -0.338 & 12 & 0 & 11 \\
should & -0.314 & 14 & 0 & 12 \\
may be & -0.300 & 20 & 2 & 18  \\
might be & -0.298 & 14 & 2 & 10  \\
must be & -0.287 & 19 & 0 & 17 \\
would & -0.284 & 16 & 0 & 15 \\
can be & -0.283 & 18 & 2 & 15 \\
would be & -0.261 & 29 & 0 & 25 \\
can & -0.208 & 16 & 1 & 13  \\
would be very & -0.186 & 8 & 0 & 6 \\[4pt]
\multicolumn{2}{l}{\bf On negative words}\\
could & 0.351 & 5 & 5 & 0\\
could be & 0.268 & 8 & 7 & 1\\
might be & 0.256 & 5 & 5 & 0\\
can be & 0.249 & 6 & 4 & 0\\
would be & 0.224 & 12 & 7 & 2\\
can & 0.200 & 5 & 4 & 0\\
may be & 0.169 & 8 & 5 & 2\\
\hline
\end{tabular}
}
\end{center}
\vspace*{-3mm}
\end{table}

In our dataset, the modal modifiers are formed by modal verbs \textit{can}, \textit{could}, \textit{should}, \textit{would}, \textit{may}, \textit{might}, and \textit{must} in combination with auxiliary verbs.
Figure~\ref{fig-modals} demonstrates the overall impact of modals on sentiment.
One can observe that on average modals have a smoothing effect on sentiment: they make negative words less negative and positive words less positive.
Words with high absolute sentiment values tend to experience the greatest shift; though, this shift is still quite small (around 0.4 points).

The effect of individual modal modifiers on positive and negative words is shown in Table~\ref{tab:modals}.
The most influential modal modifier is \textit{would have been}.
It consistently downshifts sentiment by a significant margin (about 0.5 points).
Modifiers involving modals \textit{could}, and \textit{might} also affect sentiment in a consistent and noticeable way for both positive and negative words.
Modals \textit{can} and \textit{would} form modifiers that have the smallest effect on sentiment of positive and negative words (with the exception of the modifier \textit{would have been}).

\subsection{Degree Adverbs}

As mentioned earlier, the average differences in sentiment caused by degree adverbs are quite small; many differences are negligible. 
Furthermore, these modifiers are less consistent than negators and modals; there are many degree adverbs that increase the sentiment intensity of some words from one class (positive or negative) and decrease the sentiment intensity of other words from the same class.
For example, \textit{certainly} heightens the sentiment intensity of positive word \textit{important} (by about 0.21 points), but lowers the sentiment intensity of another positive word \textit{hope} (by about 0.31 points). 
We found that the only degree adverb in our set that affects sentiment to a large extent (0.835 points) is \textit{less}; it consistently and significantly decreases the sentiment intensity of positive words.
In fact, it acts as negator and reduces the sentiment intensity of positive words to a degree similar to that of negators. 
There are a few other modifiers that consistently reduce the sentiment intensity of positive words by a significant amount: \textit{was too}, \textit{too}, \textit{probably}, \textit{fairly}, and \textit{relatively}.
Only one intensifier, \textit{highly}, consistently and significantly increase the sentiment of positive words.
The sentiment of negative words is significantly lowered by intensifiers \textit{extremely} and \textit{very very}.

\subsection{Interactive Visualization}
\label{visualization}

As part of this project, we created an interactive visualization for SCL-NMA.\footnote{www.saifmohammad.com/WebPages/SCL.html\#NMA}  
The visualization has several components that allow to investigate the effect of sentiment modifiers on individual words as well as to inspect the complete set in one scatter plot. 
The groups of modifiers are color-coded for ease of exploration. 
The full information for a phrase, including the sentiment scores of the phrase and its constituent content word, can be viewed by hovering over the point in the graph with the mouse.
The scatter plot can be filtered to show phrases that include only a particular type of the modifiers (negators, modals, or degree adverbs). 
All the components are linked together so that by clicking on a point in one component one can highlight or filter the corresponding points shown in the other components.
We hope that the users will find this visualization very helpful in exploring aspects of the data they are interested in.

\section{Conclusions}
We created a real-valued sentiment lexicon of phrases that include a variety of common sentiment modifiers such as negators, modals, and degree adverbs. 
Both phrases and their constituent content words are annotated manually using the Best--Worst Scaling technique. 
We showed that the obtained annotations are reliable---re-doing the annotation with different sets of annotators produces a very similar ranking of terms by sentiment.  
We use the annotations for the phrases to present an extensive analysis of how negators, modals, and degree adverbs impact the sentiment of other words in their scope. 
We demonstrate that these modifiers affect sentiment in complex ways so that their effect cannot be easily modeled with simple heuristics. 
In particular, we observe that the effect of a modifier is often determined not only by the type of the modifier (whether it is a negator, modal, or degree adverb) but also by the modifier word and the content word themselves. 
The created lexicon is made freely available to the research community to foster further research, especially towards automatic methods for sentiment composition and towards a better understanding of how sentiment is composed in the human brain.

\bibliography{maxdiff}
\bibliographystyle{naaclhlt2016}

\end{document}